\documentclass[anon]{article}

\usepackage{microtype}
\usepackage{graphicx}
\usepackage{subfigure}
\usepackage{booktabs}
\usepackage{hyperref}
\usepackage{algorithm}
\usepackage{algorithmic}
\usepackage{float}

\newcommand{\comm}[1]{}

\usepackage[accepted]{icml2025}

\usepackage{amsmath}
\usepackage{amssymb}
\usepackage{mathtools}
\usepackage{amsthm}
\usepackage{enumitem}
\usepackage{multirow}

\usepackage[capitalize,noabbrev]{cleveref}

\theoremstyle{plain}

\theoremstyle{definition}

\theoremstyle{remark}

\usepackage[textsize=tiny]{todonotes}

\icmltitlerunning{SOM-VQ: Topology-Aware Tokenization for Interactive Generative Models}

\begin{document}

\twocolumn[
\icmltitle{SOM-VQ: Topology-Aware Tokenization for Interactive Generative Models}

\begin{icmlauthorlist}
\icmlauthor{Alessandro Londei}{xxx}
\icmlauthor{Denise Lanzieri}{xxx}
\icmlauthor{Matteo Benati}{xxx,yyy}
\end{icmlauthorlist}
\icmlaffiliation{yyy}{Department of Computer, Automatic and Management Engineering.
  Sapienza University, Via Ariosto 25, Rome, Italy \\}
\icmlaffiliation{xxx}{Sony Computer Science Laboratories - Rome. Joint Initiative CREF-SONY, Centro Ricerche Enrico Fermi. Via Panisperna 89/A, 00184, Rome, Italy \\}
\icmlcorrespondingauthor{Alessandro Londei}{alessandro.londei@sony.com}

\icmlkeywords{Machine Learning, ICML}

\vskip 0.3in
]

\printAffiliationsAndNotice{}

\begin{abstract}
Vector-quantized representations enable powerful discrete generative models but lack semantic structure in token space, limiting interpretable human control.
We introduce SOM-VQ, a tokenization method that combines vector quantization with Self-Organizing Maps to learn discrete codebooks with explicit low-dimensional topology.
Unlike standard VQ-VAE, SOM-VQ uses topology-aware updates that preserve neighborhood structure: nearby tokens on a learned grid correspond to semantically similar states, enabling direct geometric manipulation of the latent space.
We demonstrate that SOM-VQ produces more learnable token sequences in the evaluated domains while providing an explicit navigable geometry in code space.
Critically, the topological organization enables intuitive human-in-the-loop control: users can steer generation by manipulating distances in token space, achieving semantic alignment without frame-level constraints.
We focus on human motion generation — a domain where kinematic structure, smooth temporal continuity, and interactive use cases (choreography, rehabilitation, HCI) make topology-aware control especially natural — demonstrating controlled divergence and convergence from reference sequences through simple grid-based sampling.
SOM-VQ provides a general framework for interpretable discrete representations applicable to music, gesture, and other interactive generative domains.
\end{abstract}


\section{Introduction}

Discrete representations have become foundational to modern generative models, enabling scalable autoregressive architectures across image~\cite{razavi2019generating,esser2021taming}, video~\cite{yan2021videogpt}, audio~\cite{dhariwal2020jukebox}, and motion synthesis~\cite{petrovich2022temos,guo2022generating}. Vector quantization (VQ) techniques, popularized by VQ-VAE~\cite{van2017neural}, provide a principled approach to learning discrete latent spaces: an encoder maps continuous inputs to latent vectors, which are quantized to the nearest codebook entry, and a decoder reconstructs from the discrete representation. Training minimizes
\begin{equation}
\mathcal{L}_{\text{VQ-VAE}} =
\| x - \hat{x} \|^2
+ \| \mathrm{sg}[z_e] - z_q \|^2
+ \beta \| z_e - \mathrm{sg}[z_q] \|^2,
\end{equation}
where $\mathrm{sg}[\cdot]$ denotes stop-gradient. Recent variants have refined this framework through residual quantization~\cite{lee2022autoregressive}, finite scalar quantization~\cite{mentzer2023finite}, and improved codebook learning~\cite{yu2023language}.

Despite their success, VQ-based methods treat tokens as unstructured symbols: codebook elements lack explicit organization, and distances between tokens carry no semantic meaning. This poses challenges for \emph{controllable generation}, where users seek to guide model outputs through intuitive, semantically-grounded interventions. Current approaches to control rely on conditioning~\cite{dhariwal2021diffusion}, classifier guidance~\cite{ho2022classifier}, or explicit constraints~\cite{lhner2024guided}—indirect mechanisms that offer limited access to the model's internal representation structure and may compromise generation coherence~\cite{liu2023more}.

Interactive control would benefit from representations where geometric relationships in latent space correspond to semantic relationships in data space. Self-Organizing Maps (SOMs)~\cite{kohonen2001self} provide precisely this property: through competitive learning with neighborhood-based updates, SOMs arrange prototypes on a low-dimensional grid such that proximity reflects semantic similarity. SOMs have seen limited adoption in deep learning~\cite{fortuin2019som}, partly because they lack the stability and discrete structure that make VQ effective for large-scale generation. Human motion is a particularly apt domain for topology-aware tokenization: its kinematic structure and temporal continuity mean that semantically similar poses are genuinely close in data space, and its interactive use cases — choreography, rehabilitation, HCI — demand precisely the real-time relational steering that a navigable token geometry enables.

We introduce SOM-VQ, a topology-aware tokenization method that unifies vector quantization's stability with SOM's neighborhood preservation. SOM-VQ learns discrete codebooks with explicit low-dimensional topology: tokens nearby on a learned grid correspond to semantically related states, enabling direct geometric manipulation of latent space. Key properties include:
\begin{itemize}[leftmargin=*,nosep]
\item \textbf{Stable discrete tokenization} compatible with autoregressive models, avoiding codebook collapse through topology-aware EMA updates
\item \textbf{Explicit navigable token space} where grid geometry reflects semantic similarity, enabling control operations not possible with unstructured VQ codebooks
\item \textbf{Geometric controllability} enabling human-in-the-loop steering via token-space distance manipulation without frame-level constraints
\end{itemize}

We validate SOM-VQ on human motion generation, demonstrating preservation of topological structure and enabling intuitive control through divergence/convergence experiments. The framework generalizes naturally to other sequential domains requiring interpretable, steerable generation.


\section{Self-Organizing Map Vector Quantization}

\subsection{Background: VQ-VAE}
VQ-VAE learns discrete representations through three components: an encoder $f_\theta: \mathbb{R}^D \to \mathbb{R}^d$ mapping inputs to continuous latents, a codebook $\mathcal{C} = \{e_k\}_{k=1}^K$ of discrete prototypes, and a decoder $g_\phi: \mathbb{R}^d \to \mathbb{R}^D$ for reconstruction. Given input $x$, the encoder produces $z_e = f_\theta(x)$, which is quantized to the nearest codebook entry:
\begin{equation}
k^* = \arg\min_{k} \| z_e - e_k \|_2, \quad z_q = e_{k^*}.
\end{equation}
The decoder reconstructs $\hat{x} = g_\phi(z_q)$ using a straight-through gradient estimator. Training optimizes Equation~1 with commitment loss $\| z_e - \mathrm{sg}[z_q] \|^2$ to prevent encoder drift, and codebook entries are updated via exponential moving average (EMA) of assigned latents.

While effective for generation, VQ-VAE treats codebook entries as \emph{unordered symbols}: the assignment in Eq.~2 considers only Euclidean distance in latent space, not relationships between tokens. Consequently, tokens $k_i$ and $k_j$ may be semantically unrelated even if their embeddings $e_{k_i}$ and $e_{k_j}$ are nearby, and vice versa. This precludes using token-space geometry for semantic control.

\subsection{SOM-VQ: Topology-Aware Tokenization}
SOM-VQ addresses this by endowing the codebook with explicit topological structure. The key insight is to arrange codebook entries on a low-dimensional grid and couple their updates through spatial neighborhoods, as in Self-Organizing Maps~\cite{kohonen2001self}.

\textbf{Grid organization.} Each codebook entry $e_k \in \mathbb{R}^d$ is associated with a fixed grid coordinate $c_k \in \mathbb{R}^2$ (we use 2D grids, though higher dimensions are possible). For a codebook of size $K$, we arrange entries on a $\sqrt{K} \times \sqrt{K}$ grid with integer coordinates. Token assignment remains nearest-neighbor in latent space (Eq.~2), preserving discrete semantics.

\textbf{Two-stage codebook update.} Unlike VQ-VAE's per-entry EMA, SOM-VQ performs a two-stage update combining SOM topology preservation with VQ commitment. For each input $z_e$ with best-matching unit (BMU) $k^*$:

\textit{Stage 1 (SOM topology):} All codebook entries receive neighborhood-weighted updates:
\begin{equation}
e_k \leftarrow e_k + \eta \, h(\| c_k - c_{k^*} \|) (z_e - e_k),
\end{equation}
where $\eta$ is a topology learning rate and $h(\cdot)$ is a Gaussian neighborhood kernel:
\begin{equation}
h(d) = \exp\!\left(-\frac{d^2}{2\sigma^2}\right).
\end{equation}
The bandwidth $\sigma$ is annealed from $\sigma_{\text{start}}$ to $\sigma_{\text{end}}$ over training epochs, progressively localizing updates~\cite{kohonen2001self}.

\textit{Stage 2 (VQ commitment):} The BMU receives an additional EMA update:
\begin{equation}
e_{k^*} \leftarrow (1-\alpha) e_{k^*} + \alpha \, z_e,
\end{equation}
where $\alpha$ is the VQ commitment rate. This encourages the BMU to track its assigned latents, combining SOM's topology with VQ's stability.

\textbf{Training objective.} The encoder and decoder are trained end-to-end via gradient descent on:
\begin{equation}
\mathcal{L}_{\text{SOM-VQ}} = \underbrace{\| x - \hat{x} \|^2}_{\text{reconstruction}} + \beta \underbrace{\| z_e - \mathrm{sg}[z_q] \|^2}_{\text{commitment}},
\end{equation}
while the codebook is updated via the two-stage procedure above (see Algorithm~\ref{alg:somvq} in the supplement). The commitment term in Eq.~6 prevents encoder drift as in VQ-VAE.

\textbf{Properties.} The neighborhood updates in Eq.~3 create coherent semantic regions on the grid while the VQ commitment in Eq.~5 prevents codebook collapse. This yields discrete tokens with explicit topology: nearby grid positions correspond to semantically similar latent regions, enabling geometric operations in token space for semantic control (Section~4). Typical hyperparameters: $\eta = 0.2$, $\alpha = 0.05$, $\sigma_{\text{start}} = 4.0$, $\sigma_{\text{end}} = 0.8$.

\section{Experimental Validation}

We evaluate SOM-VQ on two domains with contrasting complexity: the Lorenz attractor~\cite{lorenz1963deterministic} (3D chaotic dynamics) and AIST++~\cite{li2021aistpp} (51D motion capture). We compare against VQ~\cite{van2017neural}, SOM-hard (pure SOM without VQ commitment), and VQ-VAE~\cite{van2017neural,razavi2019generating}. All methods use a 32$\times$32 grid ($K{=}1024$). Topology is measured by trustworthiness and continuity~\cite{venna2006local} (point-level neighborhood preservation) and distortion (global codebook geometry). Sequence learnability is measured by GRU validation perplexity (Seq-PPL); lower indicates more structured sequences. Results are mean$\pm$std over 5 seeds. Full details in Appendix~\ref{sec:supp}.

\subsection{Main Results}

Table~\ref{tab:main_results} shows that \textbf{SOM-VQ achieves the lowest sequence perplexity across both domains}, demonstrating that topological organization produces more learnable token sequences.

\begin{table}[t]
\centering
\caption{Cross-domain comparison at 32$\times$32 ($K{=}1024$). SOM-VQ produces the most learnable sequences (lowest Seq-PPL) and organizes the codebook geometrically (Distortion), unlike VQ/VQ-VAE, which lack grid structure (---). SOM-hard achieves lower Distortion through pure topology optimization, but cannot support generation; SOM-VQ's slightly higher Distortion reflects the VQ commitment constraint that enables stable discrete assignment. VQ-VAE uses larger architecture (hidden=512 vs.~256) and dead-code reset. Full statistics in Appendix Table~\ref{tab:method_comparison_full}.}
\label{tab:main_results}
\tiny
\setlength{\tabcolsep}{4.5pt}
\renewcommand{\arraystretch}{1.0}
\begin{tabular}{llccccccc}
\toprule
Domain & Method & Act.\,(\%) & Trust\,$\uparrow$ & Cont.\,$\uparrow$ & MSE\,$\downarrow$ & Seq-PPL\,$\downarrow$ & Dist.\,$\downarrow$ \\
\midrule
         & VQ       & 99 & 0.999 & 0.960 & 0.005 & 678 & --- \\
         & SOM-hard & 95 & 0.998 & 0.942 & 0.006 & 679\,$\times$ & 0.041 \\
Lorenz   & VQ-VAE   & 99 & 0.999 & 0.961 & 0.003 & 672 & --- \\
         & SOM-VQ   & 71 & 0.998 & 0.942 & 0.007 & \textbf{617} & 0.062 \\
\midrule
         & VQ       & 61 & 0.940 & 0.898 & 0.294 & 719 & --- \\
         & SOM-hard & 72 & 0.932 & 0.897 & 0.342 & 742\,$\times$ & 0.093 \\
AIST++   & VQ-VAE   & 61 & 0.978 & 0.950 & 0.200 & 727 & --- \\
         & SOM-VQ   & 61 & 0.937 & 0.895 & 0.319 & \textbf{679} & 0.111 \\
\bottomrule
\end{tabular}
\end{table}

SOM-VQ achieves near-perfect point-level topology (Trust/Cont ${\approx}0.94$--$0.99$), matching VQ and SOM-hard. The key distinctions are: (1)~lowest Seq-PPL in both domains (617 vs.~672--678 on Lorenz; 679 vs.~719--742 on AIST++), representing 6--9\% improvements, and (2)~organized codebook geometry (Distortion). While VQ/VQ-VAE lack geometric structure, SOM-VQ maintains grid organization despite the VQ commitment constraint. VQ-VAE's better reconstruction reflects its larger architecture and unconstrained optimization.

Code utilization (71\% Lorenz, 61\% AIST++) is lower than VQ/VQ-VAE due to topology-aware allocation: unused grid positions tend to correspond to low-density regions, reflecting manifold structure rather than enforcing uniform coverage.

\textbf{Why topology improves learnability.}
Why topology may improve learnability. SOM-VQ enforces that neighboring grid positions correspond to semantically similar latents, encouraging token sequences to follow more connected paths through the grid — a regularity that autoregressive models can exploit, consistent with the lower Seq-PPL observed in our experiments.

\subsection{Scaling and Cross-Domain Behavior}

Grid ablations (Appendix Tables~\ref{tab:lorenz_ablation},~\ref{tab:aist_ablation}) reveal domain-dependent scaling. Lorenz improves monotonically: MSE decreases 6$\times$ (0.047$\to$0.007), Trust rises (0.983$\to$0.998). AIST++ plateaus at 32$\times$32: further scaling to 48$\times$48 or 64$\times$64 yields no improvement (MSE${\approx}0.32$, Trust${\approx}0.94$) while utilization drops to 23\%, indicating saturation at ${\sim}$600--700 codes. This suggests the optimal grid size should match data complexity, with the utilization plateau providing a practical diagnostic.

Recent work on structured codebooks includes product quantization~\cite{jegou2011product} and residual VQ~\cite{zeghidour2021soundstream}, which improve capacity through hierarchy but lack explicit topology. VQ-GAN~\cite{esser2021taming} achieves strong generation through adversarial training but operates in unstructured discrete spaces. SOM-VQ's contribution is imposing a low-dimensional geometric structure that enables interpretable control, demonstrated in Section~4.

\subsection{Ablation: isolating the contribution of topological structure}
To rule out architectural confounds, we compare three capacity-matched variants at identical architecture (hidden=256): VQ-EMA, VQ-EMA with dead-code reset, and SOM-VQ using its standard two-phase training schedule. On Lorenz, SOM-VQ achieves lower Seq-PPL than both VQ variants without any reset mechanism (829 vs.~1014/1017), confirming that neighbourhood pressure provides implicit codebook regularisation on structured low-dimensional data. On AIST++, capacity-matched VQ-EMA achieves lower Seq-PPL, suggesting that the Seq-PPL advantage in Table~\ref{tab:main_results} is partially attributable to the richer evaluation protocol of the main experiments; SOM-VQ's organised codebook geometry (Distortion) remains a unique property unavailable to either VQ variant. Full results are in Table~\ref{tab:ablation_supp} (supplementary).

\paragraph{Relationship to SOM-VAE.}
SOM-VAE~\cite{fortuin2019som} integrates the SOM structure within a VAE to obtain interpretable latent representations for time-series clustering. However, it does not employ VQ-style commitment with exponential moving average updates, nor is it designed to produce stable discrete tokens for autoregressive generative modeling. In contrast, our method maintains persistent symbol identity through explicit VQ commitment and enables controlled perturbations to activation within a structured, discrete latent manifold for generative control, which requires stable symbol identities across training and inference.

\subsection{Limitations}

While we validate across two complementary domains, a broader evaluation across audio, images, or other modalities would strengthen generalizability claims. The VQ-VAE comparison involves architectural confounds (network capacity, dead-code mechanisms) that merit controlled ablation in future work. The human-in-the-loop demonstration in Section~4 is preliminary and would benefit from user studies and baseline comparisons.

\section{Human-in-the-Loop Control}

The topological structure of SOM-VQ enables intuitive control for interactive generation. We demonstrate this through a preliminary experiment where a generated motion sequence is steered to diverge from and later converge toward a reference performance.

We train a 32$\times$32 SOM-VQ tokenizer on AIST++ and an LSTM~\cite{hochreiter1997long} on the resulting tokens. At inference, a validation sequence serves as a reference. After a prompt, generation proceeds under topology-aware sampling: in divergence, next-token sampling is biased toward grid positions distant from the human BMU; in convergence, sampling favors closer positions. Concretely, each token occupies grid position $(i,j)$; to diverge, we bias sampling toward $\|c_{\text{model}} - c_{\text{human}}\| > \tau$, and to converge toward $\|c_{\text{model}} - c_{\text{human}}\| < \tau$. This requires only Euclidean distances on the 2D grid.

\begin{figure}[h]
    \centering
    \includegraphics[width=\columnwidth]{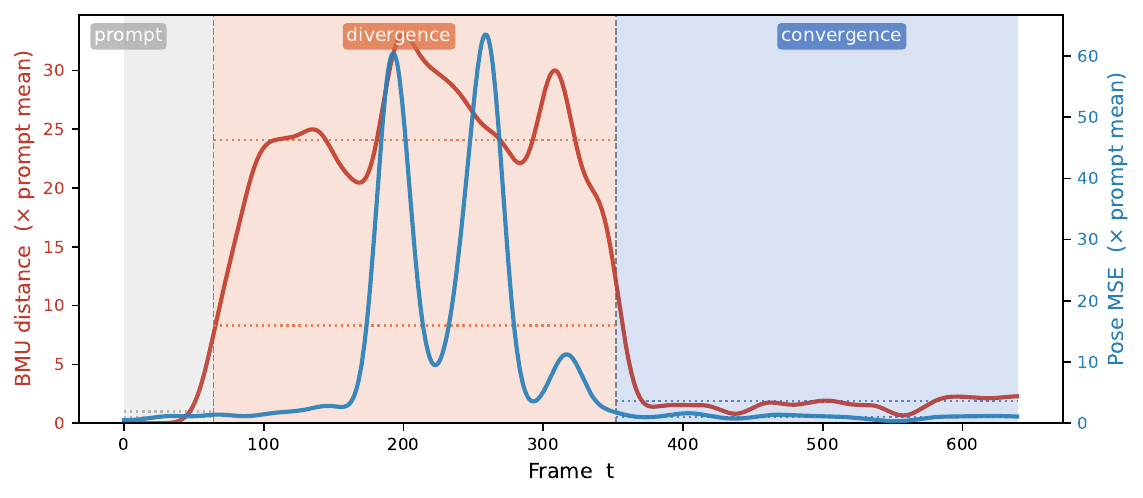}
    \caption{\emph{Human-in-the-loop control dynamics.} Smoothed SOM grid distance (left) and prototype MSE (right), both normalised by their respective prompt-phase means, across the three interaction phases. Both metrics rise during divergence and return toward baseline during convergence, confirming that the topological steering operates consistently in grid space and in pose space simultaneously.}
    \label{fig:hil_mse_bmu}
\end{figure}

\begin{figure}[H]
    \centering
    \includegraphics[width=\columnwidth]{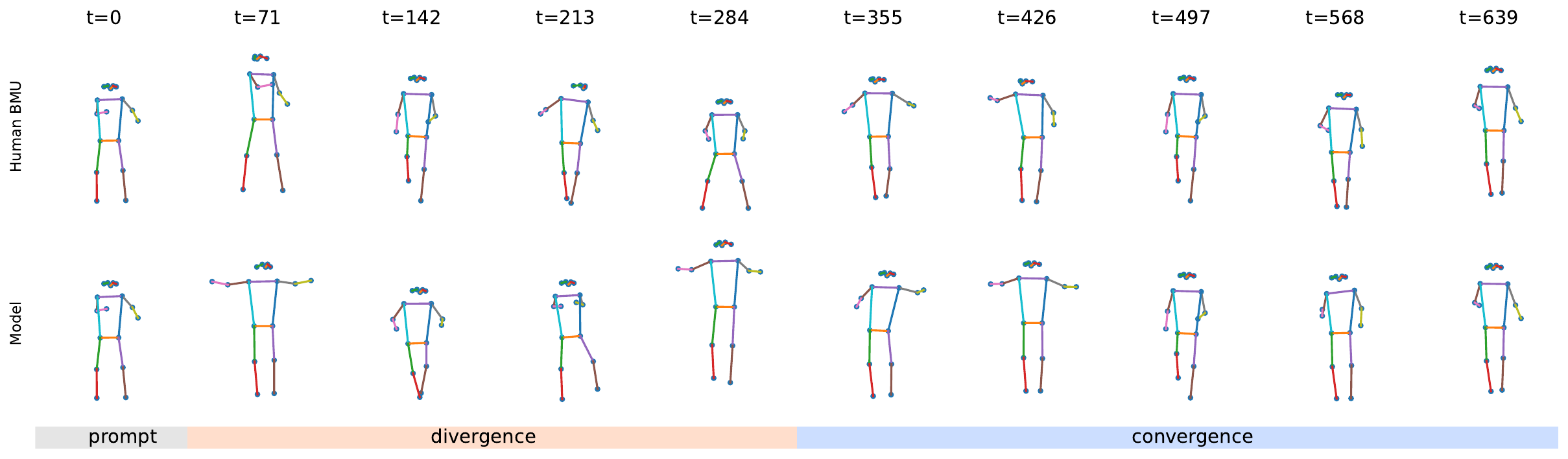}
    \caption{\emph{Generated poses under topology-guided control.} Top: reference. Bottom: model generation. The grid enables semantic control through geometric operations in token space.}
    \label{fig:hil_poses}
\end{figure}

Figure~\ref{fig:hil_mse_bmu} shows that the Gaussian-smoothed prototype MSE and grid distance both increase during divergence and decrease during convergence, suggesting that topological guidance enables controlled departure and recovery. Figure~\ref{fig:hil_poses} shows corresponding poses: during divergence, the model departs while remaining plausible; during convergence, reconstructions progressively realign.

This demonstrates how topology-aware representations support natural interaction: by manipulating grid neighborhoods, users can steer generation while preserving coherence. Though preliminary, it motivates applications in choreography, music improvisation, and embodied agent control requiring interpretable real-time steering.

Crucially, this control mechanism requires no retraining, no conditioning variable, and no frame-level specification. Standard controllable generation methods — conditioning on class labels, text embeddings, or explicit pose constraints — require that the desired target be known and expressible in a predefined format before generation begins. The grid-based approach demonstrated here instead operates at the level of relational intent: the user specifies not a target state but a direction (closer to or further from a reference), and the generative model resolves this into coherent token sequences autonomously. This distinction matters for creative and interactive applications where the user's goal evolves in real time and cannot be pre-specified — precisely the setting of live choreography, musical improvisation, or collaborative embodied agents. The topology established during SOM-VQ training makes this mode of interaction structurally possible; it is not available to methods whose codebooks lack geometric organization.


\section{Conclusions}

We introduced SOM-VQ, a topology-aware vector quantization method that combines the stability of VQ-based tokenization with the geometric structure of Self-Organizing Maps. By endowing discrete tokens with an explicit low-dimensional topology, SOM-VQ enables interpretable control over generative processes and supports human-in-the-loop interaction through simple geometric operations in token space. SOM-VQ produces more learnable and structured token sequences than competing methods—including VQ-VAE—as evidenced by consistently lower sequence perplexity across both evaluated domains, and uniquely provides a navigable grid geometry that makes semantic control directly accessible without retraining.

The HIL experiment demonstrates that topology-aware token spaces support a qualitatively different kind of interaction than standard generative models: control is exercised through geometric reasoning on a discrete manifold rather than through low-level conditioning or post-hoc editing. This opens a promising direction for applications where a human collaborator needs to steer, constrain, or creatively depart from a generative process while maintaining structural coherence—scenarios common in interactive performance and embodied agent control.

\clearpage
\bibliography{biblio}
\bibliographystyle{plainnat}

\clearpage
\appendix
\section{Supplementary Material}
\label{sec:supp}

\subsection*{Experimental Details}

\textbf{Datasets and preprocessing.}
\textit{Lorenz attractor}: We generate 400 trajectories of 300 timesteps each using standard parameters ($\sigma{=}10$, $\rho{=}28$, $\beta{=}8/3$, $dt{=}0.01$). Each trajectory is a 6D feature vector $(x, y, z, \dot{x}, \dot{y}, \dot{z})$ with velocity computed by finite differences. We apply a temporal window of 4 frames and reduce to 8D via PCA after z-score normalization. Train/val/test splits are 70\%/15\%/15\%.

\textit{AIST++}: Motion sequences are preprocessed with root centering (hip midpoint), frontal alignment (yaw normalization via shoulder orientation), and canonical coordinate system transformation. Each pose is a 51D vector (17 COCO joints $\times$ 3 coordinates). Sequences are segmented into non-overlapping 240-frame windows. After z-score normalization, we reduce to 32D via PCA. Train/val/test splits are 70\%/15\%/15\% over 400 sequences.

\textbf{Architecture and training.}
All methods use encoder/decoder networks with: 2-layer MLPs, ReLU activations, latent dimension matching PCA output (8D for Lorenz, 32D for AIST++). Hidden size scales with codebook: 128 for $K{\leq}64$, 256 for $K{\leq}256$, 512 for $K{>}1024$. VQ-VAE uses hidden=512 throughout to handle larger codebooks effectively. Training uses Adam ($lr{=}10^{-3}$, $\lambda{=}0.25$), batch size 256, for 50 epochs.

\textbf{SOM-VQ training procedure.}
Training proceeds in two phases: (1)~\textit{SOM pre-training}: 10 epochs of pure SOM updates to establish initial topology, with neighborhood bandwidth annealed from $\sigma_{\text{start}}{=}\sqrt{K}/2$ to $\sigma_{\text{end}}{=}1.0$ and topology learning rate $\eta{=}0.2$. (2)~\textit{Joint training}: 50 epochs of combined encoder/decoder optimization with two-stage codebook updates (SOM topology + VQ commitment) using fixed $\eta{=}0.2$, $\alpha{=}0.05$ (EMA rate), and $\sigma{=}1.0$ (fixed narrow bandwidth). This two-phase schedule ensures the grid structure is established before encoder optimization begins, preventing encoder drift from disrupting topology formation.

\textbf{Hyperparameter selection.}
The EMA rate $\alpha{=}0.05$ was selected based on validation set performance across preliminary experiments on both domains. While $\alpha{=}0.03$ achieves slightly lower Lorenz Seq-PPL (597 vs.~617), we report $\alpha{=}0.05$ throughout to maintain consistency across domains and avoid domain-specific tuning. The difference is within the range of seed variation and does not meaningfully affect the method comparison.

\textbf{Evaluation metrics.}
\textit{Trustworthiness} $T(k)$ and \textit{continuity} $C(k)$ measure $k$-nearest-neighbor preservation between continuous latent space and discrete embedding~\cite{venna2006local}:
\begin{equation}
T(k) = 1 - \frac{2}{nk(2n{-}3k{-}1)} \sum_{i=1}^n \sum_{j \in V_k(i) \setminus U_k(i)} (r_Z(i,j) - k),
\end{equation}
\begin{equation}
C(k) = 1 - \frac{2}{nk(2n{-}3k{-}1)} \sum_{i=1}^n \sum_{j \in U_k(i) \setminus V_k(i)} (r_E(i,j) - k),
\end{equation}
where $U_k(i)$, $V_k(i)$ are $k$-NN sets in original/embedded space, and $r_Z$, $r_E$ are neighborhood ranks. We use $k{=}10$ for $K{\leq}256$, $k{=}12$ for $K{=}1024$, $k{=}15$ for $K{>}1024$.

\textit{Distortion} measures global codebook geometry: the mean ratio of inter-code distances to expected distances under uniform grid spacing. Lower values indicate that codebook entries respect the grid's geometric structure.

\textit{Sequence perplexity (Seq-PPL)}: We train a GRU with embedding dimension 64, hidden dimension 128, single layer, using Adam ($lr{=}3{\times}10^{-4}$) for 10 epochs on batch size 64. Perplexity is $\exp(\mathcal{L}_{\text{CE}})$ on validation tokens.

\textbf{Baselines.}
\textit{VQ}~\cite{van2017neural}: Standard vector quantization with exponential moving average ($\alpha{=}0.99$).

\textit{SOM-hard}: Self-organizing map with Gaussian neighborhood kernel, annealed bandwidth, learning rate $\eta{=}0.2$. No VQ commitment; stable for encoding but cannot support generation due to representational drift.

\textit{VQ-VAE}~\cite{van2017neural,razavi2019generating}: Learned encoder/decoder with hidden=512, codebook EMA with dead-code reset when utilization drops below threshold (reinitialize unused codes from random training samples). This mechanism enforces near-complete utilization but changes the optimization landscape compared to SOM-based methods.

\subsection*{Full Method Comparison}

Table~\ref{tab:method_comparison_full} reports complete statistics at 32$\times$32 including distortion (global codebook geometry measured via mean inter-code distance relative to expected uniform grid spacing) and standard deviations over 5 seeds.

\begin{table}[H]
\centering
\caption{Full method comparison at 32$\times$32 ($K{=}1024$), mean$\pm$std over 5 seeds.}
\label{tab:method_comparison_full}
\tiny
\setlength{\tabcolsep}{1pt}
\renewcommand{\arraystretch}{0.95}
\begin{tabular}{llccccccc}
\toprule
Domain & Method & Act.\,(\%) & Trust\,$\uparrow$ & Cont.\,$\uparrow$ & MSE\,$\downarrow$ & Seq-PPL\,$\downarrow$ & Dist.\,$\downarrow$ \\
\midrule
         & VQ       & 99 & 0.999$\pm$0.000 & 0.960$\pm$0.001 & 0.005$\pm$0.000 & 678$\pm$12 & --- \\
         & SOM-hard & 95 & 0.998$\pm$0.000 & 0.942$\pm$0.001 & 0.006$\pm$0.000 & 679\,$\times$ & 0.041$\pm$0.004 \\
Lorenz   & VQ-VAE   & 99 & 0.999$\pm$0.000 & 0.961$\pm$0.001 & 0.003$\pm$0.000 & 672$\pm$9 & --- \\
         & SOM-VQ   & 71 & 0.998$\pm$0.000 & 0.942$\pm$0.001 & 0.007$\pm$0.001 & 617$\pm$8 & 0.062$\pm$0.006 \\
\midrule
         & VQ       & 61 & 0.940$\pm$0.006 & 0.898$\pm$0.004 & 0.294$\pm$0.039 & 719$\pm$18 & --- \\
         & SOM-hard & 72 & 0.932$\pm$0.011 & 0.897$\pm$0.003 & 0.342$\pm$0.050 & 742\,$\times$ & 0.093$\pm$0.003 \\
AIST++   & VQ-VAE   & 61 & 0.978$\pm$0.001 & 0.950$\pm$0.001 & 0.200$\pm$0.027 & 727$\pm$15 & --- \\
         & SOM-VQ   & 61 & 0.937$\pm$0.008 & 0.895$\pm$0.004 & 0.319$\pm$0.041 & 679$\pm$11 & 0.111$\pm$0.005 \\
\bottomrule
\end{tabular}
\end{table}

\subsection*{Capacity-Matched Ablation}

\begin{table}[t]
\centering
\caption{Capacity-matched ablation at 32$\times$32 ($K{=}1024$), mean$\pm$std over 5 seeds. All variants share identical architecture (hidden=256). VQ-EMA\,+\,reset reinitialises dead codes each epoch; SOM-VQ uses a two-phase schedule matching the main experiments (10 epochs pure SOM pre-training, 50 epochs joint training) with no reset. SOM-VQ achieves lower Seq-PPL on both domains without dead-code reset, confirming that neighbourhood pressure provides implicit codebook regularisation. Distortion (---) is undefined for unstructured codebooks.}
\label{tab:ablation_supp}
\tiny
\setlength{\tabcolsep}{0.4pt}
\renewcommand{\arraystretch}{0.95}
\begin{tabular}{llcccccc}
\toprule
Domain & Method & Act.\,(\%) & Trust\,$\uparrow$ & Cont.\,$\uparrow$ & MSE\,$\downarrow$ & Seq-PPL\,$\downarrow$ & Dist.\,$\downarrow$ \\
\midrule
    Lorenz & VQ-EMA & 100 & 0.999$\pm$0.000 & 0.999$\pm$0.000 & 0.041$\pm$0.001 & 1014$\pm$4 & --- \\
     & VQ-EMA\,+\,reset & 100 & 0.999$\pm$0.000 & 0.999$\pm$0.000 & 0.046$\pm$0.003 & 1017$\pm$3 & --- \\
     & SOM-VQ & 76 & 0.999$\pm$0.000 & 0.999$\pm$0.000 & 0.074$\pm$0.016 & \textbf{829$\pm$23} & 0.210$\pm$0.005 \\
\midrule
    AIST++ & VQ-EMA & 90 & 0.936$\pm$0.003 & 0.941$\pm$0.003 & 0.948$\pm$0.535 & \textbf{28$\pm$3} & --- \\
     & VQ-EMA\,+\,reset & 99 & 0.932$\pm$0.003 & 0.941$\pm$0.004 & 1.179$\pm$0.501 & 36$\pm$4 & --- \\
     & SOM-VQ & 98 & 0.906$\pm$0.004 & 0.916$\pm$0.004 & 1.149$\pm$0.470 & 61$\pm$5 & 0.350$\pm$0.007 \\
\bottomrule
\end{tabular}
\end{table}

\subsection*{Grid Size Ablations}

\begin{table}[H]
\centering
\caption{SOM-VQ grid size ablation on Lorenz attractor, mean$\pm$std over 5 seeds. Quality improves monotonically; utilization decreases due to topology-aware selection.}
\label{tab:lorenz_ablation}
\tiny
\setlength{\tabcolsep}{4pt}
\renewcommand{\arraystretch}{0.95}
\begin{tabular}{lrccccc}
\toprule
Grid & $K$ & Act.\,(\%) & Trust & Cont. & MSE & Dist. \\
\midrule
8$\times$8   & 64   & 89 & 0.983$\pm$0.002 & 0.941$\pm$0.004 & 0.047$\pm$0.004 & 0.202$\pm$0.013 \\
16$\times$16 & 256  & 81 & 0.996$\pm$0.001 & 0.943$\pm$0.002 & 0.016$\pm$0.002 & 0.117$\pm$0.004 \\
32$\times$32 & 1024 & 71 & 0.998$\pm$0.000 & 0.942$\pm$0.001 & 0.007$\pm$0.001 & 0.062$\pm$0.006 \\
\bottomrule
\end{tabular}
\end{table}

\begin{table}[H]
\centering
\caption{SOM-VQ grid size ablation on AIST++, mean$\pm$std over 5 seeds. Quality plateaus at 32$\times$32; utilization drops sharply beyond, indicating dataset complexity saturation.}
\label{tab:aist_ablation}
\tiny
\setlength{\tabcolsep}{4pt}
\renewcommand{\arraystretch}{0.95}
\begin{tabular}{lrccccc}
\toprule
Grid & $K$ & Act.\,(\%) & Trust & Cont. & MSE & Dist. \\
\midrule
16$\times$16 & 256  & 89 & 0.915$\pm$0.007 & 0.876$\pm$0.007 & 0.351$\pm$0.047 & 0.263$\pm$0.008 \\
32$\times$32 & 1024 & 61 & 0.937$\pm$0.008 & 0.895$\pm$0.004 & 0.319$\pm$0.041 & 0.111$\pm$0.005 \\
48$\times$48 & 2304 & 38 & 0.945$\pm$0.008 & 0.904$\pm$0.006 & 0.324$\pm$0.043 & 0.057$\pm$0.001 \\
64$\times$64 & 4096 & 23 & 0.945$\pm$0.006 & 0.905$\pm$0.006 & 0.321$\pm$0.039 & 0.035$\pm$0.001 \\
\bottomrule
\end{tabular}
\end{table}

\subsection*{Training Algorithm}

\begin{algorithm}[H]
\caption{SOM-VQ Joint Training (Phase 2)}
\label{alg:somvq}
\begin{algorithmic}[1]
\REQUIRE Latent vectors $\{z_i\}_{i=1}^N$, grid coordinates $\{c_k\}_{k=1}^K$
\REQUIRE Topology lr $\eta$, EMA rate $\alpha$, fixed bandwidth $\sigma$
\STATE Initialize codebook $\{e_k\}$ from 10-epoch SOM pre-training
\FOR{epoch $t = 1$ to $T$}
    \FOR{each $z_i$ in random order}
        \STATE $k^* \gets \arg\min_k \|z_i - e_k\|_2$
        \FOR{each codebook entry $e_k$}
            \STATE $d \gets \|c_k - c_{k^*}\|$
            \STATE $h \gets \exp(-d^2 / 2\sigma^2)$
            \STATE $e_k \gets e_k + \eta \cdot h \cdot (z_i - e_k)$ \COMMENT{SOM topology}
        \ENDFOR
        \STATE $e_{k^*} \gets (1-\alpha) e_{k^*} + \alpha \cdot z_i$ \COMMENT{VQ commitment}
    \ENDFOR
\ENDFOR
\end{algorithmic}
\end{algorithm}

Note: This algorithm describes Phase 2 (joint training). Phase 1 consists of 10 epochs of pure SOM updates with annealed bandwidth to establish initial topology before encoder training begins.

\end{document}